\documentclass[letterpaper, 10 pt, conference]{ieeeconf}  

\IEEEoverridecommandlockouts                              

\title{\LARGE \bf
A Joint Modeling of Vision-Language-Action for Target-oriented Grasping in Clutter
}

\author{Kechun Xu, Shuqi Zhao, Zhongxiang Zhou, Zizhang Li, Huaijin Pi, Yue Wang, Rong Xiong
\thanks{{This work was supported by the National Nature Science Foundation of China under Grant 62173293}. Kechun Xu, Shuqi Zhao, Zhongxiang Zhou, Zizhang Li, Huaijin Pi, Yue Wang, Rong Xiong are with Zhejiang University,
Hangzhou, China. Corresponding author,{\tt\small wangyue@iipc.zju.edu.cn}.}
}

\usepackage{color}
\usepackage{amsmath}
\usepackage{graphicx}
\usepackage{subfigure}
\usepackage{makecell}
\usepackage{booktabs}
\usepackage{gensymb}
\usepackage{textcomp}
\usepackage{cite}
\usepackage{url}
\usepackage[colorlinks,
            allcolors=blue,
            linkcolor=blue,
            anchorcolor=blue,
            citecolor=blue]{hyperref}

\begin{document}

\maketitle
\thispagestyle{empty}
\pagestyle{empty}

\begin{abstract}

We focus on the task of language-conditioned grasping in clutter, in which a robot is supposed to grasp the target object based on a language instruction. Previous works separately conduct visual grounding to localize the target object, and generate a grasp for that object. However, these works require object labels or visual attributes for grounding, which calls for handcrafted rules in planner and restricts the range of language instructions. In this paper, we propose to jointly model vision, language and action with object-centric representation. Our method is applicable under more flexible language instructions, and not limited by visual grounding error. Besides, by utilizing the powerful priors from the pre-trained multi-modal model and grasp model, sample efficiency is effectively improved and the sim2real problem is relived without additional data for transfer. A series of experiments carried out in simulation and real world indicate that our method can achieve better task success rate by less times of motion under more flexible language instructions. Moreover, our method is capable of generalizing better to scenarios with unseen objects and language instructions. Our code is available at {\href{https://github.com/xukechun/Vision-Language-Grasping}{https://github.com/xukechun/Vision-Language-Grasping}}.

\end{abstract}

\section{Introduction}

Target-oriented grasping in clutter is an essential task for robotic manipulation, and has been explored for decades \cite{zeng2022robotic, murali20206, fang2018multi, jang2017end}. Common practices to specify the target object are to provide a target object image \cite{zeng2022robotic, sun2021gater}, or collect a set of demonstrations \cite{laskey2016robot}. But providing such information is often infeasible for a user, especially in open-world applications, such as tabletop rearrangement over open-vocabulary objects.

An intuitive idea is to replace visual command with natural language to assign the target object. Recent progress in pre-trained large language models \cite{devlin2018bert, brown2020language} and multi-modal models \cite{radford2021learning} makes this idea practical and promising. In  \cite{ahn2022can, huang2022inner, shridhar2022cliport, zheng2022vlmbench}, the robot is able to learn policies for multi-task ({\it e.g.} pick-and-place and room rearrangement) by grounding language instructions as task goals with these pre-trained models. However, these works learn raw image based policy with lots of simulated demonstration data. In addition, one has to address sim2real gap to deploy raw image based policy in real world. The cluttered scene may further increase the data requirement and sim2real difficulty.

Instead, object-centric representation endows the robot with awareness of objects in the scene, which enables easier scene understanding, thus more efficient for grasping tasks. In a similar vein, another line of works \cite{hatori2018interactively, shridhar2018interactive, yang2022interactive, zhang2021invigorate, goodwin2022semantically} decouples the task of language-conditioned target-oriented grasping into two separate steps: language visual grounding to figure out the target object and grasp planning to pick up that object. In \cite{shridhar2018interactive, yang2022interactive, zhang2021invigorate}, the robot generates object-centric representations by object bounding boxes, grounds semantics with object labels and visual attributes, and selects an action by a rule-based planner. However, these handcrafted attributes and rules restrict the generalization of language instructions. Furthermore, error in visual grounding results, and disturbance from other objects in the cluttered scene might degenerate the successful grasping, or call for more complex handcrafted rules in planning.

\begin{figure}[t]
  \centering
  \includegraphics[width=0.88\linewidth]{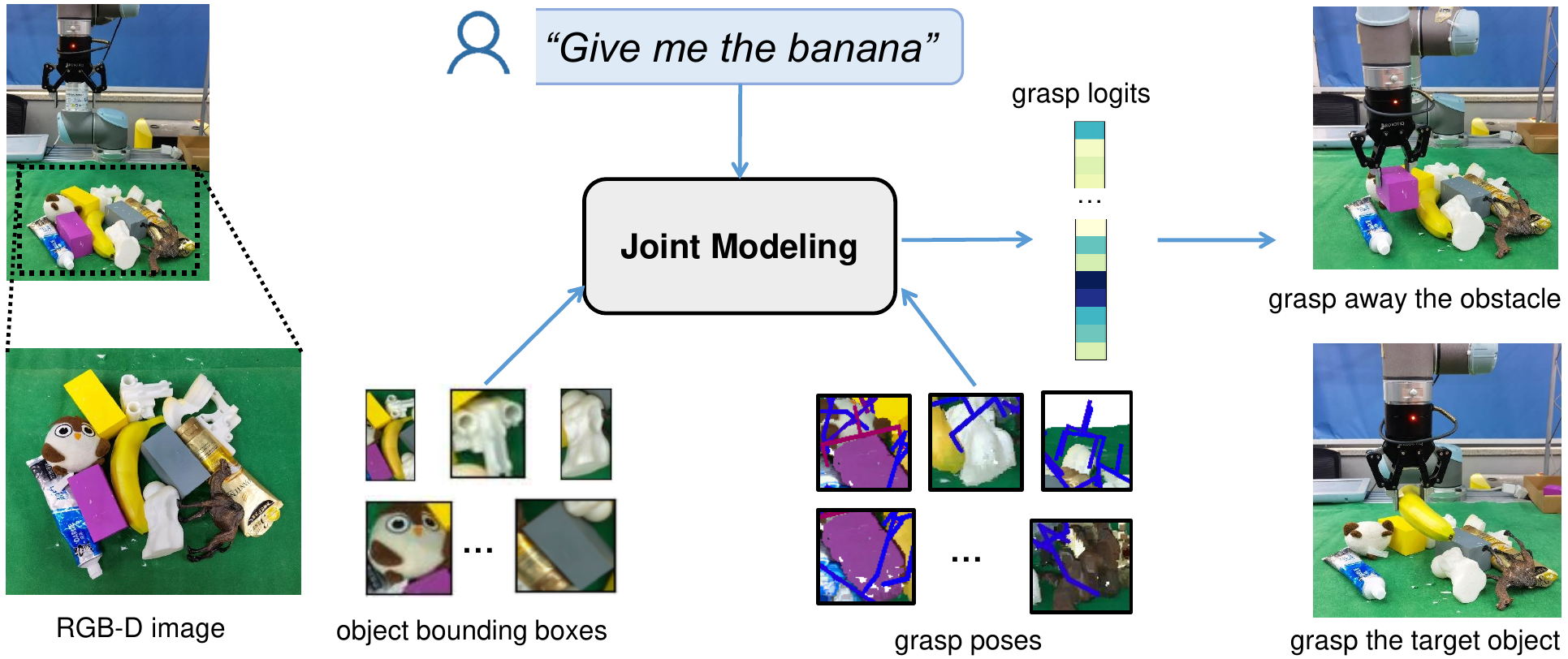}
  \vspace{-0.3cm}
  \caption{An example scenario of language-conditioned grasping in clutter. Our method jointly models vision, language and action with object-centric representation ({\it i.e.} object bounding box), and conducts a sequence of grasps to grasp away obstacles, finally achieving the target object.}
  \label{fig:teaser}
  \vspace{-0.7cm}
\end{figure}

In this paper, we set to investigate an intermediate solution between the previous two lines of works. We propose to jointly model vision, language and action with object-centric representations as input, and grasp poses as output (Fig. \ref{fig:teaser}). Specifically, we utilize the pre-trained visual-language model CLIP~\cite{radford2021learning} to encode object-centric bounding boxes and the language instruction to visual-language features, and deploy the pre-trained graspnet \cite{fang2020graspnet} to generate a set of grasp poses, which are encoded to spatial features. We apply a transformer to build the cross-attention features with grasp features as queries, visual-position features as keys and visual-language features as values. Then the cross-attention features are fed into the policy network for action decision. As a whole, our policy is learned through model-free deep reinforcement learning in simulator, which predicts a sequence of grasps to grasp away objects hindering the successful grasping, and finally achieve the goal. 

Compared to the first line of works, leveraging pre-trained models with object-centric representation provides powerful priors and object awareness for policy learning, thus improving the sample efficiency. Also, considering that the processes of generating object-centric representation and grasp poses from raw images cause sim2real gap, we deploy pre-trained models learned with massive simulation and real data for these two processes, relieving this problem. Compared to the second line, our policy directly takes the multi-modal features for grasp selection, which eliminates the explicit design of object labels or visual attributes in planner or language instructions, and avoids relying on visual grounding. Consequently, our method is able to handle more flexible language instructions, and achieve better language generalization.
Experiments conducted in simulation and real-world environments demonstrate that our system can achieve higher task success rate by less steps in language-conditioned target-oriented grasping in clutter. We also show that our method is capable of performing better in scenarios with unseen objects and language instructions, validating the generalization. To summarize, our main contributions are:

\begin{itemize}
    \item We propose a joint modeling of vision, language and action for the task of target-oriented grasping in clutter through model-free deep reinforcement learning.
    \item We utilize the priors from the pre-trained visual-language model and grasp model to relieve sim2real gap without additional data for transfer, and improve the sample efficiency.
    \item The learned system is evaluated on a series of scenarios with seen and unseen objects and language instructions in both simulated and real-world settings, of which the results validate the effectiveness and generalization.
\end{itemize}

\section{Related Works}

\subsection{Target-oriented Grasping in Clutter}

Robotic grasping in clutter has been a topic of interest in manipulation for decades. Recent works \cite{pinto2016supersizing, mahler2017learning, kalashnikov2018scalable, ten2018using, mahler2017dex} have achieved significant improvements in grasping general objects in clutter, and some of them \cite{fang2020graspnet, sundermeyer2021contact, wang2021graspness, son2022grasping} are capable of generating 6DoF grasp proposals of a clutter of every objects. Another line of works studies target-oriented grasping by assigning an image of the target object \cite{sun2021gater}, or providing a demonstration \cite{laskey2016robot}. Recent works \cite{zeng2022robotic, murali20206, fang2018multi, jang2017end} step forward to explore target-oriented grasping in clutter. \cite{zeng2022robotic, jang2017end} achieve the target object grasping by grasping other obstacles away. \cite{kiatos2019robust, kurenkov2020visuomotor, yang2020deep, xu2021efficient, huang2021visual} conduct some pre-grasping actions such as pushing to singulate the target object. However, these works require an image of the target object, or assume that a target object mask in the scene is given, which is often infeasible, especially in open-world applications, such as tabletop rearrangement. Instead, language instructions are more flexible in open-world applications \cite{shridhar2022cliport, stepputtis2020language}.

\subsection{Language-conditioned Manipulation}

Because of the convenience of natural language, language-conditioned manipulation has become a promising research topic in recent years. \cite{rao2018learning, ito2022integrated, chen2021joint} study grasp detection based on language instructions in scattered scenes. \cite{misra2016tell} explores to ground language to some long-horizon manipulation tasks, but with instructions limitation. \cite{stepputtis2020language} introduces language as a flexible goal specification with objects of simple shape in scattered scenes. Thanks to the develop of language models \cite{devlin2018bert, brown2020language} and multi-modal models \cite{radford2021learning}, recent works \cite{ahn2022can, huang2022inner, shridhar2022cliport, zheng2022vlmbench, stengel2022guiding} are able to ground more flexible language instructions into long-horizon manipulation. However, such works consume lots of demonstrations to learn raw image based policy for a good performance, and take plenty of training steps for convergence. In addition, one has to address sim2real gap to deploy raw image based policy in real world. Other works \cite{shridhar2018interactive, goodwin2022semantically, yang2022interactive, zhang2021invigorate, hatori2018interactively, ahn2018interactive} focus on language-conditioned grasping by separately visual grounding and grasp planning with object-centric representations. \cite{shridhar2018interactive, goodwin2022semantically, yang2022interactive} ground object bounding boxes in the scene with pre-defined visual attributes, or object labels, which calls for handcrafted design in planner, and limits the range of language instructions. To overcome visual grounding error and language ambiguity, they need to interact with human by asking questions with pre-defined templates. In this line of work, language-conditioned grasping in clutter is much less explored except that \cite{hatori2018interactively, zhang2021invigorate}. \cite{hatori2018interactively} conducts direct visual grounding by calculating the cosine similarity between text feature and object features, which cannot handle visual grounding error and visual occlusion. Analogous to \cite{yang2022interactive}, \cite{zhang2021invigorate} trains on RefCOCO dataset \cite{yu2016modeling} and requires object labels for grounding.

In this paper, we also focus on the task of language-conditioned grasping in clutter. There are several differences between our methods and the above target-oriented grasping works. Instead of grounding a matching score of each object bounding box from language instructions, we utilize the fusion feature of visual and language from CLIP to learn a grasp policy without handcrafted grounding rules, thus applicable to more flexible language instructions. Also, by jointly modeling vision, language and action and conducting cross attention with grasp queries, our system focuses on the ability of all grasp actions in the scene, rather than relying on the visual grounding to select an object.

\section{Methods}
\subsection{System Overview}

\begin{figure*}[t]
  \centering
  \includegraphics[width=0.88\textwidth]{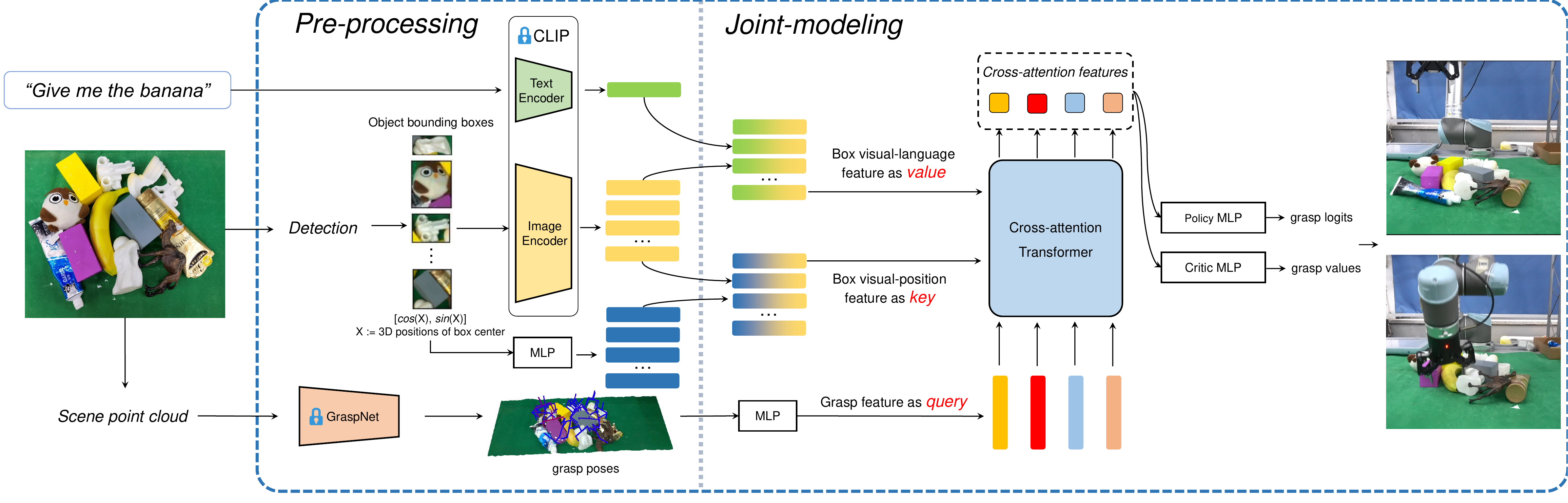}
  \vspace{-0.4cm}
  \caption{{\bf System Overview}. Given a language instruction, and object bounding boxes from detection module, our system pre-processes visual-language input by CLIP \cite{radford2021learning}, and generates grasp poses by graspnet \cite{fang2020graspnet} (see Sec.~\ref{method:preprocess}). And we jointly model vision-language-action by a cross-attention transformer, of which outputs are fed into the policy and critic MLPs to generate logits and values of all grasps (see Sec.~\ref{method:joint}).}
  \label{fig:overview}
  \vspace{-0.6cm}
\end{figure*}

As shown in Fig. \ref{fig:overview}, given a language instruction, and a set of object bounding boxes from the detection module, our system generates visual features of all bounding boxes and language feature by pre-trained CLIP \cite{radford2021learning} image encoder and text encoder. Each bounding box visual feature is fused with language feature to generate box visual-language feature, while fused with the embedding of bounding box center position to generate box visual-position feature. We utilize the pre-trained graspnet \cite{fang2020graspnet} to yield a set of grasp poses which are encoded to spatial features. Given these feature embeddings, we propose to jointly model vision, language and grasp action by a cross-attention transformer, which takes grasp features as queries, box visual-position features as keys, box visual-language features as values, and outputs the cross-attention features. Then the policy and critic MLPs take as input the cross-attention features, and output the logits and values of all grasps respectively. At inference time, our policy conducts a sequence of grasp poses to grasp away the obstacles and finally catch the target object.
\subsection{Pre-processing with Pre-trained Models}
\label{method:preprocess}
{\bf Object-centric Representation.} In this paper, we obtain object bounding boxes for object-centric representation, which are fed into pre-trained models for pre-processing.

{\bf Visual-language Model.} We propose to yield visual and language features through CLIP \cite{radford2021learning}, which aligns image and language features by training on millions of image-text pairs. Instead of directly applying CLIP on the raw image of a clutter of objects, we propose to process object crops in the scene. Given a raw image with a clutter of $M$ objects, $N$ bounding boxes are extracted from detection models, and $N$ object crops are obtained by cropping the raw image with corresponding bounding boxes. Note that $M$ is not equal to $N$ in most cases due to object occlusion in clutter and the detection uncertainty. Then $N$ object crops are fed into the CLIP image encoder and the language instruction is fed into the CLIP text encoder to get $N$ bounding box visual features and a language feature in CLIP image-text alignment domain.

{\bf Grasp Model.} We adopt the pre-trained model of graspnet \cite{fang2020graspnet} to get a set of grasp poses of the whole scene. Graspnet is a 6DoF grasp detector taking as input a scene point cloud, and predicting $K$ 6DoF grasp poses.

Thanks to pre-trained vision-language and grasp model, the sample efficiency is effectively improved. Also, utilizing CLIP \cite{radford2021learning} and graspnet \cite{fang2020graspnet} trained with massive simulation and real data for pre-processing relieves the sim2real gap.

\subsection{Vision-language-action Joint Modeling}
\label{method:joint}
{\bf Vision-language-action Cross-attention.} Instead of separately conducting visual grounding and selecting a best-matched object to grasp, we propose to jointly model vision, language and action by transformer’s attention mechanism \cite{vaswani2017attention}: $\text{Attention}(Q, K, V)=\text{softmax}\left({Q K^T}\right) V$, where $Q, K, V$ donate query, key and value. Specifically, a cross-attention transformer is adapted to conduct cross attention among multi-modal information, and outputs fusion features. We fuse $N$ box visual features with the language feature by element-wise product respectively and get $N$ box visual-language features accordingly. Also, we generate $K$ grasp pose features by MLP. Considering that our goal of joint modeling is to get the fusion vision-language-action features, and to map grasp features into the CLIP image-text feature domain, we set $K$ grasp pose features as queries and $N$ box visual-language features as values. To combine spatial features ({\it i.e.} grasp features) and visual-language features, we further utilize the box visual-position features as keys. To be specific, 3D positions of the centers of bounding boxes are projected into a nonlinear space (positional embedding as in \cite{mildenhall2020nerf}), followed by a MLP to encode the box position embeddings. Then $N$ box position embeddings are fused with $N$ box visual features by adding. After vision-language-action cross attention, $K$ cross-attention features will be obtained.

{\bf Policy and Critic.} In this work, our system is trained through deep reinforcement learning with discrete SAC \cite{haarnoja2018soft, christodoulou2019soft}. We consider the cross-attention features at step $t$ as state $s_t$, the grasp pose selection at step $t$ as action $a_t$. Then our policy is defined as $\pi(a_t|s_t)$, and critic outputs Q function $Q(s_t, a_t)$. The policy and critic MLPs both take $K$ cross-attention features as input, and output logits and q values of $K$ grasp poses. Note that our policy and critic are capable of processing variable number of actions by parallel processing $K$ cross-attention features, which corresponds to $K$ grasps.

{\bf Language Instructions.} By jointly modeling vision-language-action and conducting cross attention with grasp queries, our system focuses on the ability of grasp actions without requirement of object labels or visual attributes for grounding and handcrafted rules for planning. This formulation enables our system to adapt to more flexible language instructions without assigning concrete object labels. Our system is trained with language templates such as ``Give me the \{keyword\}'', ``Get something to \{keyword\}'', where \{keyword\} can be a concrete label ({\it e.g. }banana), or a general label ({\it e.g. }fruit), or the attribute of color ({\it e.g. }red), shape ({\it e.g. }round), or even object function ({\it e.g. }hold other things). Although some of the language instructions are ambiguous, our policy can learn from experience by trial and error.
\vspace{-0.1cm}
\subsection{Curriculum Learning Schedule}
\label{method-curriculum}

Directly training a policy in clutter might make the agent struggle with the trade-off between grasping away other obstacles and focusing on the target object. Considering that our final goal is to grasp the target object, we design a two-stage curriculum learning schedule to ease the training:

{\bf Stage I. Scattered Scene Training.} In this stage, we train our policy in a scattered scene with $a$ objects randomly dropped in the workspace, which aims at guiding the robot to focus on grasping the target object. The number of grasps is limited up to 5, and once the target object is successfully grasped, the episode ends. The reward function is defined as:
\vspace{-0.12cm}
\begin{equation}
\label{eqn:grasp reward}
R_{g}=\left\{\begin{array}{ll}
2, & \text { if\;successfully\;grasping\;the\;target\;object} \\
-1, & \text { otherwise }
\end{array}\right.
\nonumber 
\vspace{-0.12cm}
\end{equation}

{\bf Stage II. Cluttered Scene Training.} For this stage, our policy is trained in cluttered environments with $b$ objects randomly dropped in the workspace, where the tight packing or occlusion degenerates the successful grasp of the target object. For each episode, the robot is supposed to grasp away obstacles hindering the grasping of the target object, and finally grasp the target object. The number of grasps is limited up to 8. In this stage, we encourage the robot to grasp the obstacle objects near the target object. Therefore, we implement less punishment if the object is nearer to the target, and we model the reward function as follows:
\vspace{-0.12cm}
\begin{equation}
\label{eqn:grasp reward}
R_{g}=\left\{\begin{array}{ll}
2, & \text { if\;successfully\;grasping\;the\;target\;object} \\
-\frac{dist}{dist_{max}}, & \text { if\;successfully\;grasping\;other\;objects}\\
-1, & \text { otherwise }\\
\end{array}\right.
\nonumber
\vspace{-0.12cm}
\end{equation}
where $dist$ is the 3D position distance between the grasped object and the target object. If there are more than one target objects, it is the minimize distance from all target objects. $dist_{max}$ is the diagonal length of the workspace. Dividing $dist$ with $dist_{max}$ normalizes the value within $[-1,0)$.

{\bf CLIP Grounding Prior.} To optimise CLIP to guide the policy to select grasp actions of the target object, we further explore a CLIP-guided loss. We ground $N$ visual-language pairs into $N$ probabilities with CLIP, which represent the corresponding visual-language matching scores, and multiply the probabilities with the $N\times K$ box-grasping mapping matrix, which assigns grasps to boxes by 3D position distance. In this way, we can get $K$ mapped grasp probabilities as the grounding prior for the policy. Then a guided loss ({\it i.e.} Kullback-Leibler divergence) can be added for the policy training. Details can be found in Appendix \cite{appendix_arxiv}.


\subsection{Implementation Details}
We train our system with five language templates: ``Give me the \{keyword\}'', ``I need a \{keyword\}'', ``Grasp a \{keyword\} object'', ``I want a \{keyword\} object'', ``Get something to \{keyword\}'' with 66 object models and 36 language keywords, which are categorized into four types: label, general label, shape or color, function. Network parameters of CLIP and graspnet will be fixed during training. We train the first stage with $a=8$ objects and 500 episodes, and  the second stage with $b=15$ objects and 1500 episodes. If using CLIP-guided loss, the policy loss and the loss of the cross-attention module will be added with the CLIP-guided loss described in Sec.~\ref{method-curriculum} for the former 800 episodes. Other training details of can be accessed in Appendix \cite{appendix_arxiv}.

\section{Experiments}
In this section, we carry out a series of experiments to evaluate our system. The goals of the experiments are: 1) to demonstrate that our policy is effective for the task of language-conditioned grasping in clutter; 2) to evaluate the generalization performance of our policy on unseen objects and language instructions. 3) to indicate the sample efficiency of our method. 4) to test whether our system can successfully transfer from simulation to the real world. We compare the performance of our system to the following baselines: 

{\bf CLIPORT-Grasp} is a method which learns actions from raw image. It is a variant of CLIPORT \cite{shridhar2022cliport}, and utilizes CLIP to ground semantic concepts from raw image. Grasp actions are also learned from scratch by predicting pixel-wise affordance on raw image. To be specific, we use the attention stream of \cite{shridhar2022cliport} for grasping.

{\bf CLIP Grounding} is a semantic grounding approach which grounds all object bounding boxes with language instruction by CLIP \cite{radford2021learning}. Specifically, it computes similarities between the CLIP visual features of boxes and the language feature, and selects the bounding box with the highest similarity as the target. Then the policy randomly samples a grasp pose corresponding to that box by the box-grasp mapping matrix described in Sec. \ref{method-curriculum}. If there is not grasp for that box, the policy will randomly select a grasp pose.

{\bf Raw Image} is a policy which uses CLIP to process the raw image, and utilizes graspnet to yield grasp poses. This baseline is to demonstrate the performance of the policy without object-centric representation, but with grasp priors. The network architecture is the same as ours.

{\bf Raw Image Grids} is a policy similar to {\bf Raw Image}. Instead, it uses grid crops of the raw image as CLIP inputs. 

For fair comparison, all the baselines are trained with the same schedule with discrete SAC as ours except {\bf CLIP Grounding}, which directly deploys CLIP for grounding.

\begin{figure}[t]
  \centering
  \includegraphics[width=0.98\linewidth]{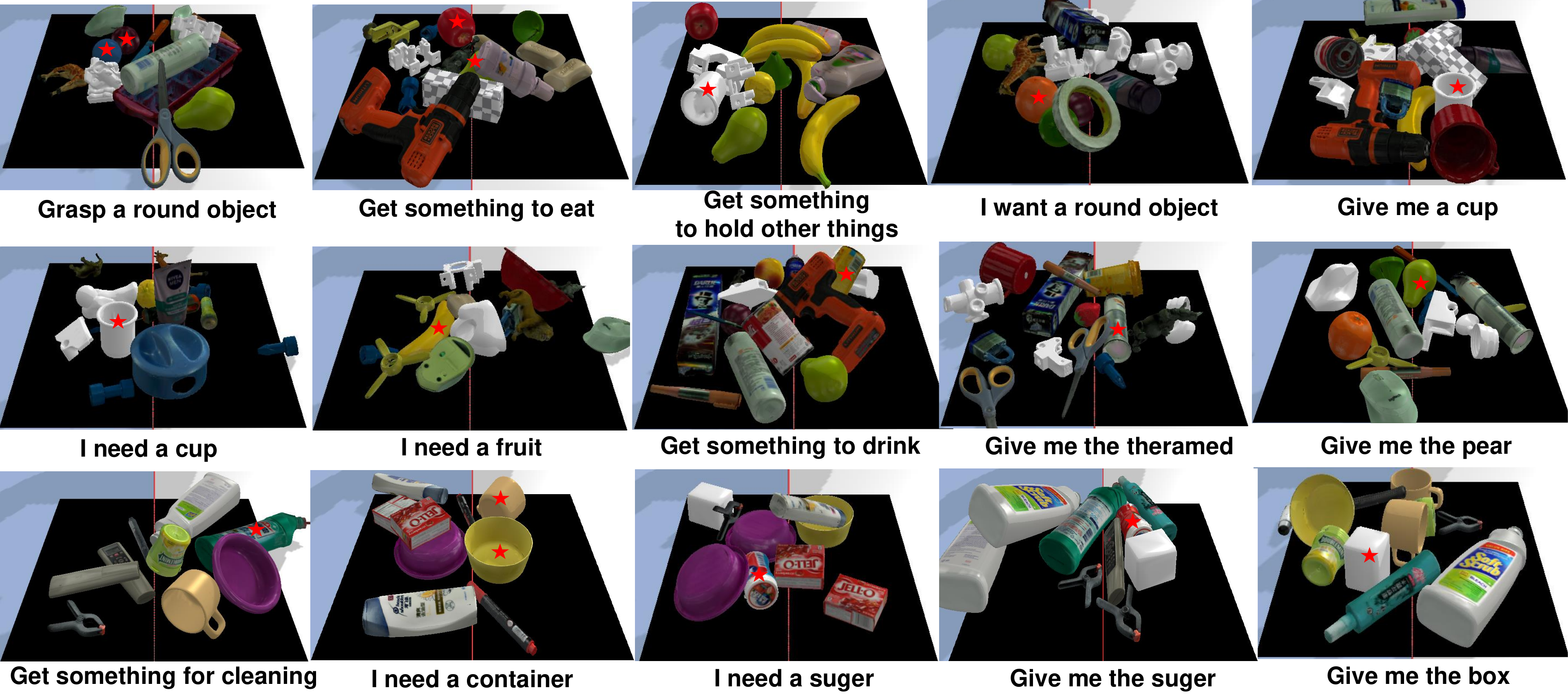}
  \vspace{-0.35cm}
  \caption{Test cases in simulation. The top two rows demonstrate the arrangements with seen objects, while the bottom row shows the arrangements with unseen objects. The target objects are labeled with stars.}
  \label{fig:simulation_case}
  \vspace{-0.6cm}
\end{figure}


\subsection{Evaluation Metrics}
We evaluate the methods with a series of test cases. Each test contains $i=15$ runs measured with 2 metrics:
\begin{itemize}
    \item {\bf Task Success Rate}: the average percentage of task success rate over $i=15$ test runs. If the robot picks up the target object within $j=8$ action attempts in a run, then the task is considered successful and completed.
    \item {\bf Motion Number}: the average motion number per task completion.
\end{itemize}

\subsection{Simulation Experiments}
Our simulation environment is built in PyBullet \cite{coumans2021}, which involves a UR5 arm, a ROBOTIQ-85 gripper, and a camera of Intel RealSense L515. 


\begin{table}[t]

\caption{{SIMULATION RESULTS ON ALL ARRANGEMENTS}}
\label{table:1}
\vspace{-0.4cm}
\begin{center}
\begin{tabular}{p{3cm}|p{0.75cm}|p{0.75cm}|p{0.75cm}|p{0.75cm}}
\hline
\makecell[c]{Method} & \multicolumn{2}{c|}{Task Success} & \multicolumn{2}{c}{Motion Number} \\
\hline
\makecell[c]{Arrangement} & \makecell[c]{seen} & {unseen} & \makecell[c]{seen}  & {unseen} \\
\hline
\makecell[c]{CLIPORT-Grasp} & \makecell[c]{--} & \makecell[c]{--} & \makecell[c]{--} & \makecell[c]{--} \\
\makecell[c]{CLIP Grounding} & \makecell[c]{59.3} & \makecell[c]{64.0} & \makecell[c]{4.80} & \makecell[c]{4.15} \\
\makecell[c]{Raw Image} & \makecell[c]{48.1} & \makecell[c]{45.3} & \makecell[c]{4.35} & \makecell[c]{5.37} \\
\makecell[c]{Raw Image Grids} & \makecell[c]{60.7} & \makecell[c]{70.7} & \makecell[c]{4.35} & \makecell[c]{{\color{red}  \bf 3.49}}\\
\makecell[c]{Ours} & \makecell[c]{\color{blue} \bf 74.3} & \makecell[c]{{\color{red} \bf  78.7}} & \makecell[c]{\color{blue} \bf 4.11} & \makecell[c]{3.98} \\
{Ours w/ CLIP-guided loss} & \makecell[c]{{\color{red} \bf 77.3}}  & \makecell[c]{\color{blue} \bf 74.7} & \makecell[c]{{\color{red} \bf 3.74}}& \makecell[c]{\color{blue} \bf 3.65}\\
\hline
\end{tabular}
\end{center}
\vspace{-0.5cm}
\end{table}

\begin{figure}[t]
  \centering
  \includegraphics[width=0.82\linewidth]{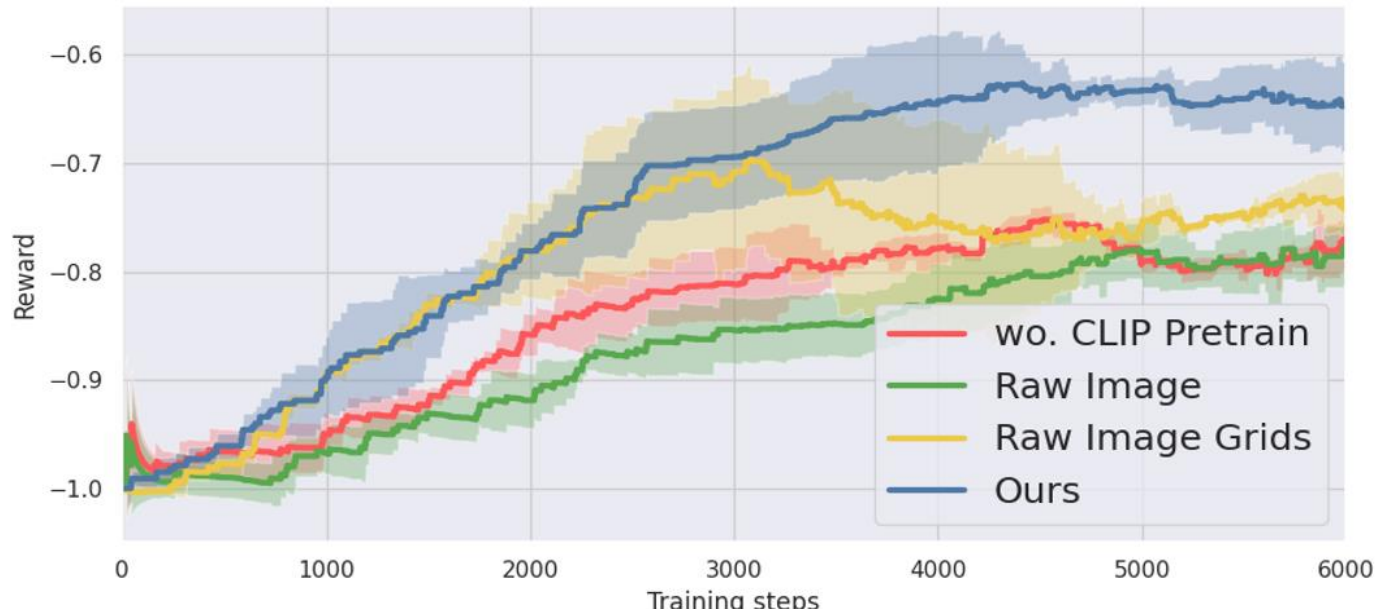}
  \vspace{-0.35cm}
  \caption{Compared training performance of our method and three methods.}
  \label{fig:training_performance}
  \vspace{-0.7cm}
\end{figure}
\begin{figure}[t]
  \centering
  \subfigure{
        \begin{minipage}[b]{0.49\linewidth}
        \centering
            \includegraphics[width=\linewidth]{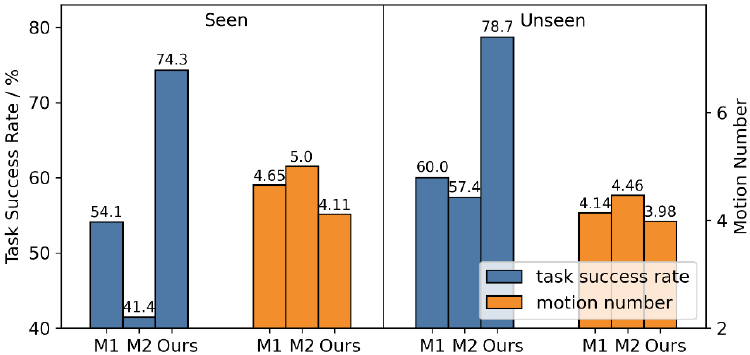}
            \label{fig:ablation-clip}
        \end{minipage}
    }
    \subfigure{
        \begin{minipage}[b]{0.42\linewidth}
        \centering
            \includegraphics[width=\linewidth]{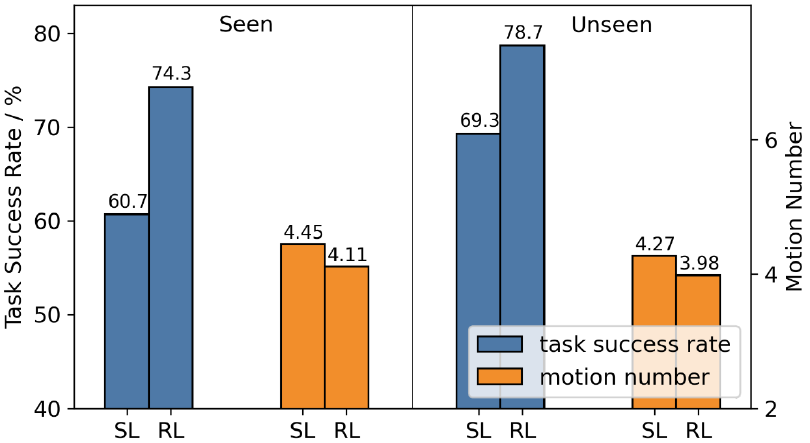}
            \label{fig:ablation-sl}
        \end{minipage}
    }
    \vspace{-0.7cm}
    \caption{Ablation on all arrangements for CLIP~({left}) and RL vs. SL~({right}).}
    \label{fig:ablation}
  \vspace{-0.3cm}
\end{figure}

{\bf Comparisons to Baselines.} We conduct test experiments with a series of test cases, including arrangements with seen objects and unseen objects (Fig. \ref{fig:simulation_case}). Arrangements with seen objects involve 10 cases with adversarial clutter where the robot might need to grasp away other obstacles for target object grasping. Compared results reported in Table \ref{table:1} demonstrate that our method outperforms all baselines across all metrics. {\bf CLIPORT-Grasp} fails in all cases, which suggests that learning grasp actions from scratch and grounding semantics from raw image might fail to work with just 2000 episodes training. Also, {\bf CLIPORT-Grasp} conducts top-down grasping, which might fail in clutter in many cases. Instead, with object-centric representation and pre-trained grasp model, {\bf CLIP Grounding} can achieve around 60$\%$ task success rate, but with the highest motion number. This might due to the fact that {\bf CLIP Grounding} tends to focus on the best-matched object and keep grasping it, and ignores the possibility of grasping away other obstacles, which is effective for grasping in clutter. Also, it can not handle the error of the grounding results. {\bf Raw Image} shows the second worst task success rate, and always cannot fix the target object in the cluttered scenarios. Instead, {\bf Raw Image Grids} gets better performance than {\bf Raw Image}. This performance gain mainly comes from the grid crops, which help the policy easier to focus on the target zone. However, there is little object-level information in grids, which might account for its unsatisfactory performance. By leveraging object-centric bounding boxes and jointly modeling vision, language and action, our policy can focus on the target object without limiting by the grounding error. Also, {\bf Ours w/ CLIP-guided loss} can achieve better performance, especially in motion number, indicating that the CLIP prior helps the policy to be more concerned with the target object.

Results of 5 test cases with unseen objects are also shown in Table \ref{table:1}. We observe that our method is capable of generalizing to unseen objects with similar language instructions, and achieves the best task success rate with a bit more motion number. Although {\bf Raw Imgae Grids} achieves the least motion number, it does not prove its high efficiency since its task success rate is lower. It's worth noting that {\bf Ours w/ CLIP-guided loss} reports lower task success rate than ours. This might due to the fact that this policy is influenced by CLIP prior to a greater extent. Although it conducts more target-driven grasps in some cases, as a trade-off, it is also affected by the CLIP grounding error.

\begin{figure}[t]
  \centering
  \includegraphics[width=0.94\linewidth]{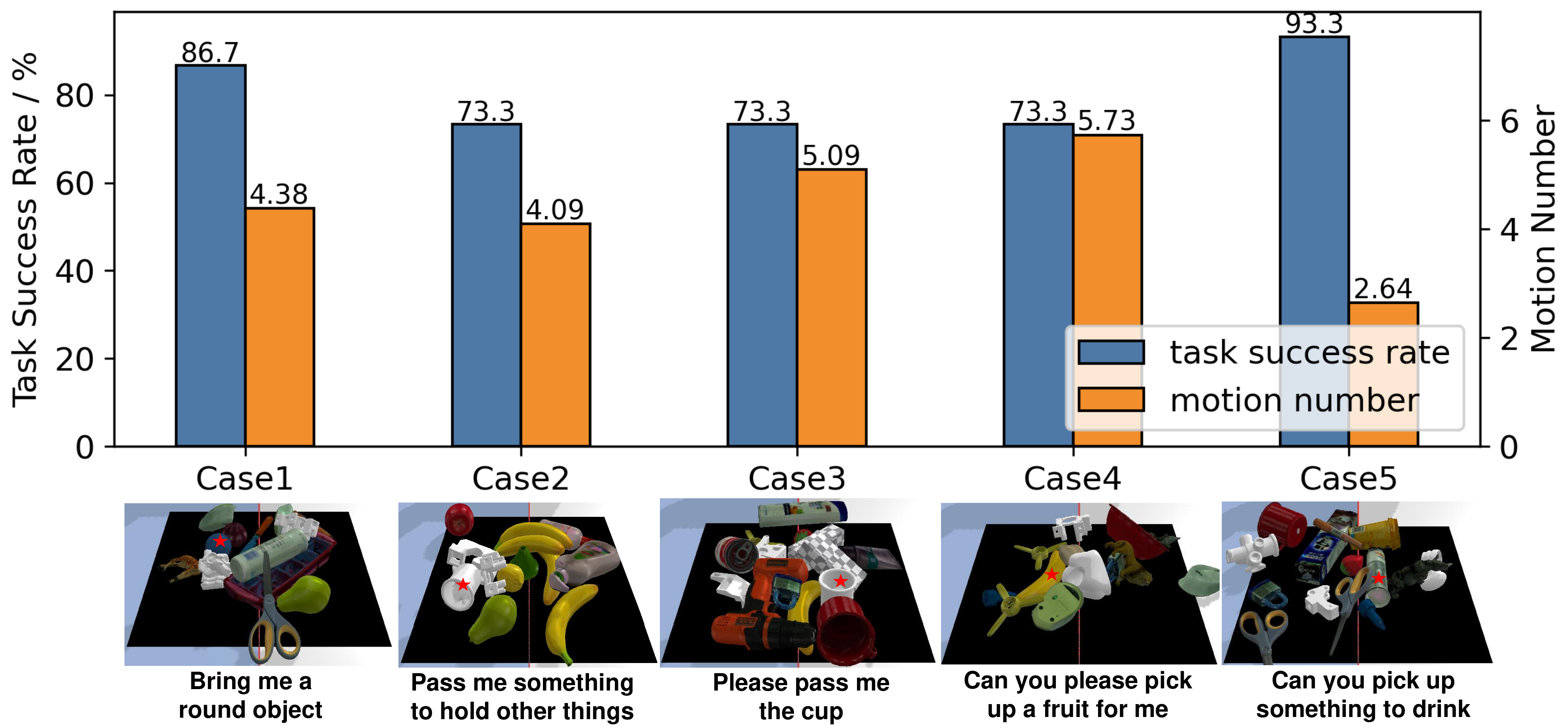}
  \vspace{-0.3cm}
  \caption{Testing performance on unseen language templates.}
  \label{fig:unseen-language}
  \vspace{-0.7cm}
\end{figure}

{\bf Ablation Studies.} We compare our methods with a series of ablation methods to test: 1) whether our techniques can improve sample efficiency; 2) whether CLIP brings benefits to our policy; 3) whether our system gains more by learning with trial and error.

To demonstrate 1), we compare our training efficiency with {\bf Ours w/o CLIP-pretrain}, {\bf Raw Image} and {\bf Raw Image Grids} to indicate whether our method benefits from pre-trained visual-language model and object-centric representation. We present the reward curves versus training steps in Fig. \ref{fig:training_performance}. It can be seen that our policy can improve performance at a faster pace and achieve a higher reward. This suggests that the CLIP prior and object-centric representation help our policy to capture vision-language information, thus improving the sample efficiency. {\bf Raw Image Grids} shows better training performance than {\bf Raw Image} by using image crops rather than raw image with a clutter of objects. It is interesting to note that {\bf Raw Image Grids} can keep a similar pace as our method in 2k$\sim$3k steps, but fails to improve later. This shows that {\bf Raw Image Grids} can handle scattered scenes (Stage I), but struggles to capture useful information in clutter with simple crops representation.


For 2), we make comparisons between our policy and two ablation methods: {\bf M1: Ours w/o CLIP-pretrain} which trains CLIP from scratch, and {\bf M2: Ours w/ RN50-BERT} which encodes image and text by ImageNet-trained ResNet50~\cite{he2016deep} and BERT~\cite{devlin2018bert}. Testing results are shown in Fig. \ref{fig:ablation}~(left). It's obvious that without CLIP prior, {\bf Ours w/o CLIP-pretrain} performs worse than ours, which suggests that pre-trained CLIP is a powerful prior benefiting the policy training. {\bf Ours w/ RN50-BERT} achieves the worst performance, indicating the unfeasibility of simply combining two pre-trained single-modal models. Instead, CLIP aligns visual feature and language feature into a common feature space, thus simplifying the fusion of vision and language and easing the policy learning.

Also, we additionally train our policy with supervised learning, which is trained with the same amount of successful action sequences as reinforcement learning, and testing results (Fig. \ref{fig:ablation}~(right)) show the advantage of training by trial and error. This might be because the supervised policy is unable to be competent for scenarios differing much from those in training set. And the training action sequences are not enough to cover the testing domain. Instead, training with RL by trial and error implicitly learns a value function for actions, which might bring better generalization performance.

\begin{figure}[t]
  \centering
  \includegraphics[width=0.95\linewidth]{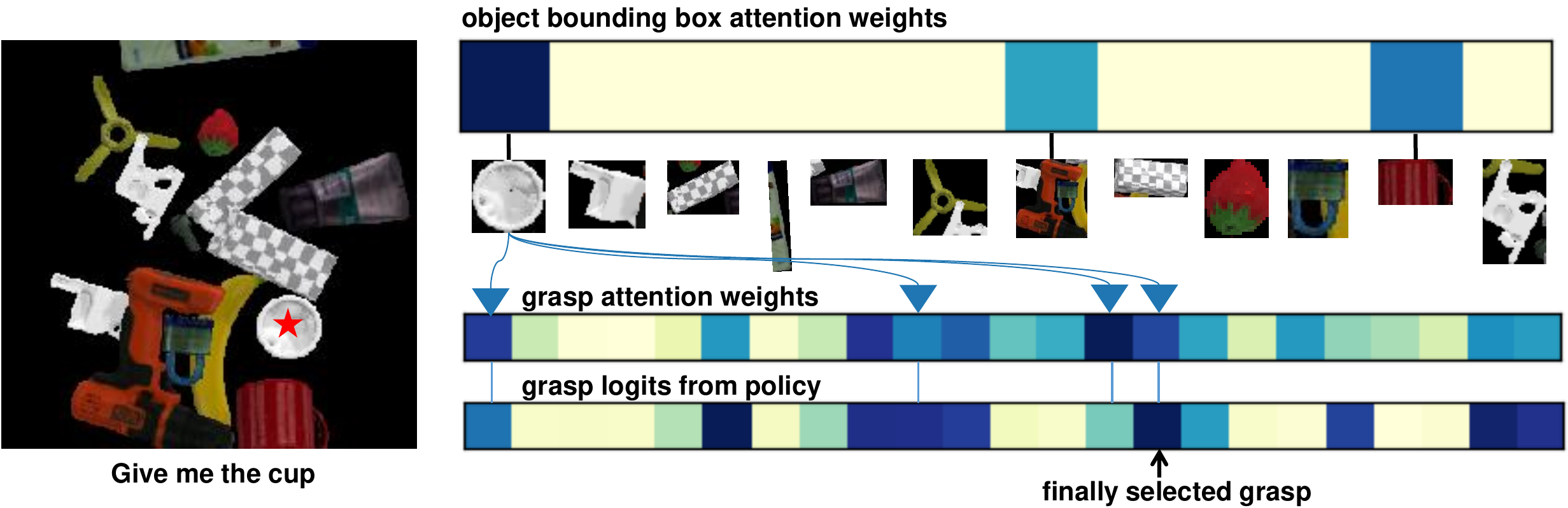}
  \vspace{-0.4cm}
  \caption{A case study to show the cross attention weights. Given a scenario where the target object is labeled with a star, we show the attention weights of all bounding boxes (top) and grasps of the box with the highest attention weight (middle). The bottom row shows the grasp logits outputted by policy. Note that grasp poses are assigned to boxes by the box-grasp mapping matrix in Sec. \ref{method-curriculum}. We mark the grasp poses of the box with triangles, and the selected grasp of the policy with an arrow.}
  \label{fig:attn_map}
  \vspace{-0.3cm}
\end{figure}

\begin{figure}[t]
  \centering
  \includegraphics[width=0.95\linewidth]{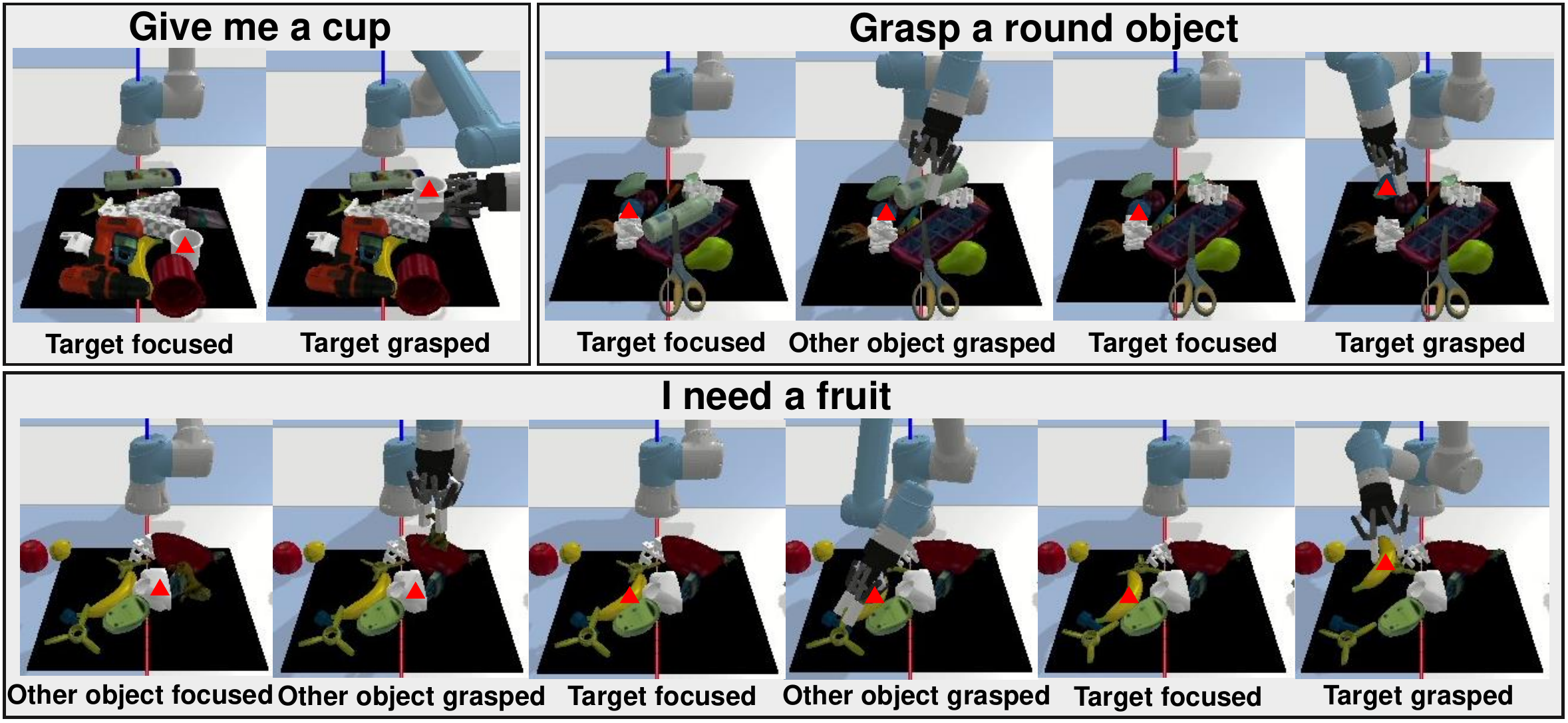}
  \vspace{-0.3cm}
  \caption{Example testing sequences in simulation of 3 cases. Objects with the highest attention weights are marked with triangles.}
  \label{fig:simultion-sequence}
  \vspace{-0.7cm}
\end{figure}

{\bf Novel Language Templates.} Furthermore, we test our policy with some novel language templates. As shown in Fig. \ref{fig:unseen-language}, we evaluate our policy with 5 scenarios containing seen objects, but under unseen language templates such as ``Please pass me the $\{$keyword$\}$'', ``Can you please pick up a $\{$keyword$\}$ for me''. Results suggest that our policy is applicable for some unseen language templates, which further confirms our claim on language generalization.

{\bf Case Studies.} Fig. \ref{fig:attn_map} visualizes the cross attention weights of a test case. We can see that the target object can be focused with the highest attention weight, followed by two surrounding objects, which means that the cross-attention mechanism is able to catch the target object with spatial information. Also, grasp poses for the target object can achieve high attention weights. Taking the cross-attention features, our policy further adjusts the logits distribution of grasp poses and draws more attention to the target. Also, we report three typical cases in Fig. \ref{fig:simultion-sequence}. In the first case, the target object can be focused and grasped successfully, which is the most ideal situation. In more situations (the second case), the policy chooses to grasp the surrounding object for better grasping of the target. When failed to focus on the target at the beginning (the third case), our policy is capable of adjusting by exploring with a sequence of grasp actions. 

\begin{figure}[t]
  \centering
  \includegraphics[width=0.9\linewidth]{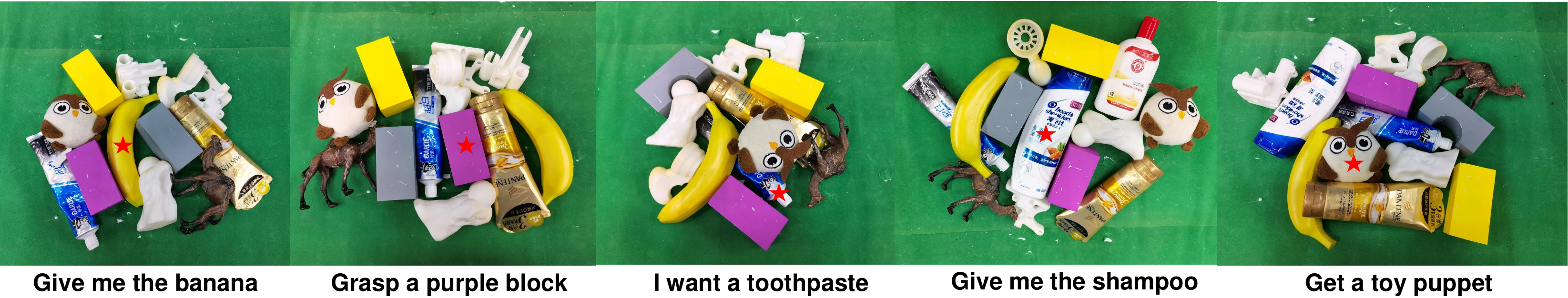}
  \vspace{-0.3cm}
  \caption{Test cases in real world. Target objects are labeled with stars.}
  \label{fig:real_case}
  \vspace{-0.2cm}
\end{figure}

\begin{table}[t]
\scriptsize
\caption{{REAL-WORLD RESULTS ON TEST ARRANGEMENTS}}
\label{table:real}
\vspace{-0.4cm}
\begin{center}
\begin{tabular}{p{2.5cm}|p{1.8cm}|p{1.9cm}}
\hline
\makecell[c]{Method} &\makecell[c]{Task Success} & {Motion Number} \\
\hline
\makecell[c]{CLIP Grounding} & \makecell[c]{56.0} & \makecell[c]{{\color{red} \bf 4.84}} \\
\makecell[c]{Raw Image Grids} & \makecell[c]{48.0} & \makecell[c]{5.33} \\
\makecell[c]{Ours} & \makecell[c]{{\color{red} \bf 72.0}} & \makecell[c]{5.13} \\
\hline
\end{tabular}
\end{center}
\vspace{-0.8cm}
\end{table}
\subsection{Real-world Experiments}
In this section, we evaluate our system in real-world settings. Our real-world platform involves a UR5 robot arm with a ROBOTIQ-85 gripper, and an Intel RealSense L515 capturing RGB-D images of resolution 1280 $\times$ 720. Object bounding boxes are generated by a pre-trained open-set detection model (details in Appendix \cite{appendix_arxiv}), and test cases are shown in Fig. \ref{fig:real_case}. We compare our methods with {\bf Raw Image Grids} and {\bf CLIP Grounding} which have demonstrated better performances than others in simulation experiments. Note that these methods are transferred from simulation to the real world without any retraining. Test results are reported in Table \ref{table:real}. Overall, our policy shows much better performance in task success rate, and is capable of generalizing to unseen target objects ({\it i.e.} purple block, toothpaste and toy puppet). Although {\bf CLIP Grounding} costs the least motion number, it shows a task success rate low at 50$\sim$60$\%$. For case 1 and 2, where CLIP grounds with high accuracy, {\bf CLIP Grounding} can focus on the target grasping with less steps. However, when the target is occluded (case 3), {\bf CLIP Grounding} frequently fails to figure out the target. {\bf Raw Image Grids} shows the worst performance in real-world cases, which tends to select grasps in a zone for the same case. Also, it demonstrates a significant performance degradation, which might be because that the sim2real gap ({\it e.g.} background) affects the crop grids input. Instead, by using bounding boxes, the influence of background can be decreased.
\vspace{-0.1cm}
\section{Conclusions}
In this work, we study the task of language-conditioned grasping in clutter. Instead of following the cascaded pipeline of visual grounding and grasp detection, we propose to jointly model vision, language and action in a union view, which is not limited by visual grounding error and is applicable with more flexible language instructions. We leverage pre-trained CLIP, a multi-modal model aligning visual and language into a common feature space, and utilize 6DoF graspnet to generate grasps in the scene, which effectively improves the sample efficiency. We train our policy through deep reinforcement learning and evaluate it with a series of experiments in simulation and real world, which indicates that our method can achieve better task success rate with less motion number compared to other methods. Moreover, our method shows better generalization in scenarios with unseen objects and language instructions. It is worth mentioning that our method can also handle some novel language templates. More discussions can be accessed in Appendix \cite{appendix_arxiv}.

\clearpage
\bibliographystyle{IEEEtran}
\bibliography{IEEEabrv,ref}

\begin{thebibliography}{10}
\providecommand{\url}[1]{#1}
\csname url@samestyle\endcsname
\providecommand{\newblock}{\relax}
\providecommand{\bibinfo}[2]{#2}
\providecommand{\BIBentrySTDinterwordspacing}{\spaceskip=0pt\relax}
\providecommand{\BIBentryALTinterwordstretchfactor}{4}
\providecommand{\BIBentryALTinterwordspacing}{\spaceskip=\fontdimen2\font plus
\BIBentryALTinterwordstretchfactor\fontdimen3\font minus
  \fontdimen4\font\relax}
\providecommand{\BIBforeignlanguage}[2]{{%
\expandafter\ifx\csname l@#1\endcsname\relax
\typeout{** WARNING: IEEEtran.bst: No hyphenation pattern has been}%
\typeout{** loaded for the language `#1'. Using the pattern for}%
\typeout{** the default language instead.}%
\else
\language=\csname l@#1\endcsname
\fi
#2}}
\providecommand{\BIBdecl}{\relax}
\BIBdecl

\bibitem{zeng2022robotic}
A.~Zeng, S.~Song, K.-T. Yu, E.~Donlon, F.~R. Hogan, M.~Bauza, D.~Ma, O.~Taylor,
  M.~Liu, E.~Romo \emph{et~al.}, ``Robotic pick-and-place of novel objects in
  clutter with multi-affordance grasping and cross-domain image matching,''
  \emph{The International Journal of Robotics Research}, vol.~41, no.~7, pp.
  690--705, 2022.

\bibitem{murali20206}
A.~Murali, A.~Mousavian, C.~Eppner, C.~Paxton, and D.~Fox, ``6-dof grasping for
  target-driven object manipulation in clutter,'' in \emph{2020 IEEE
  International Conference on Robotics and Automation (ICRA)}.\hskip 1em plus
  0.5em minus 0.4em\relax IEEE, 2020, pp. 6232--6238.

\bibitem{fang2018multi}
K.~Fang, Y.~Bai, S.~Hinterstoisser, S.~Savarese, and M.~Kalakrishnan,
  ``Multi-task domain adaptation for deep learning of instance grasping from
  simulation,'' in \emph{2018 IEEE International Conference on Robotics and
  Automation (ICRA)}.\hskip 1em plus 0.5em minus 0.4em\relax IEEE, 2018, pp.
  3516--3523.

\bibitem{jang2017end}
E.~Jang, S.~Vijayanarasimhan, P.~Pastor, J.~Ibarz, and S.~Levine, ``End-to-end
  learning of semantic grasping,'' in \emph{Conference on Robot
  Learning}.\hskip 1em plus 0.5em minus 0.4em\relax PMLR, 2017, pp. 119--132.

\bibitem{sun2021gater}
M.~Sun and Y.~Gao, ``Gater: Learning grasp-action-target embeddings and
  relations for task-specific grasping,'' \emph{IEEE Robotics and Automation
  Letters}, vol.~7, no.~1, pp. 618--625, 2021.

\bibitem{laskey2016robot}
M.~Laskey, J.~Lee, C.~Chuck, D.~Gealy, W.~Hsieh, F.~T. Pokorny, A.~D. Dragan,
  and K.~Goldberg, ``Robot grasping in clutter: Using a hierarchy of
  supervisors for learning from demonstrations,'' in \emph{2016 IEEE
  international conference on automation science and engineering (CASE)}.\hskip
  1em plus 0.5em minus 0.4em\relax IEEE, 2016, pp. 827--834.

\bibitem{devlin2018bert}
J.~Devlin, M.-W. Chang, K.~Lee, and K.~Toutanova, ``Bert: Pre-training of deep
  bidirectional transformers for language understanding,'' \emph{arXiv preprint
  arXiv:1810.04805}, 2018.

\bibitem{brown2020language}
T.~Brown, B.~Mann, N.~Ryder, M.~Subbiah, J.~D. Kaplan, P.~Dhariwal,
  A.~Neelakantan, P.~Shyam, G.~Sastry, A.~Askell \emph{et~al.}, ``Language
  models are few-shot learners,'' \emph{Advances in neural information
  processing systems}, vol.~33, pp. 1877--1901, 2020.

\bibitem{radford2021learning}
A.~Radford, J.~W. Kim, C.~Hallacy, A.~Ramesh, G.~Goh, S.~Agarwal, G.~Sastry,
  A.~Askell, P.~Mishkin, J.~Clark \emph{et~al.}, ``Learning transferable visual
  models from natural language supervision,'' in \emph{International Conference
  on Machine Learning}.\hskip 1em plus 0.5em minus 0.4em\relax PMLR, 2021, pp.
  8748--8763.

\bibitem{ahn2022can}
M.~Ahn, A.~Brohan, N.~Brown, Y.~Chebotar, O.~Cortes, B.~David, C.~Finn,
  K.~Gopalakrishnan, K.~Hausman, A.~Herzog \emph{et~al.}, ``Do as i can, not as
  i say: Grounding language in robotic affordances,'' \emph{arXiv preprint
  arXiv:2204.01691}, 2022.

\bibitem{huang2022inner}
W.~Huang, F.~Xia, T.~Xiao, H.~Chan, J.~Liang, P.~Florence, A.~Zeng, J.~Tompson,
  I.~Mordatch, Y.~Chebotar \emph{et~al.}, ``Inner monologue: Embodied reasoning
  through planning with language models,'' \emph{arXiv preprint
  arXiv:2207.05608}, 2022.

\bibitem{shridhar2022cliport}
M.~Shridhar, L.~Manuelli, and D.~Fox, ``Cliport: What and where pathways for
  robotic manipulation,'' in \emph{Conference on Robot Learning}.\hskip 1em
  plus 0.5em minus 0.4em\relax PMLR, 2022, pp. 894--906.

\bibitem{zheng2022vlmbench}
K.~Zheng, X.~Chen, O.~C. Jenkins, and X.~E. Wang, ``Vlmbench: A compositional
  benchmark for vision-and-language manipulation,'' \emph{arXiv preprint
  arXiv:2206.08522}, 2022.

\bibitem{hatori2018interactively}
J.~Hatori, Y.~Kikuchi, S.~Kobayashi, K.~Takahashi, Y.~Tsuboi, Y.~Unno, W.~Ko,
  and J.~Tan, ``Interactively picking real-world objects with unconstrained
  spoken language instructions,'' in \emph{2018 IEEE International Conference
  on Robotics and Automation (ICRA)}.\hskip 1em plus 0.5em minus 0.4em\relax
  IEEE, 2018, pp. 3774--3781.

\bibitem{shridhar2018interactive}
M.~Shridhar and D.~Hsu, ``Interactive visual grounding of referring expressions
  for human-robot interaction,'' \emph{arXiv preprint arXiv:1806.03831}, 2018.

\bibitem{yang2022interactive}
Y.~Yang, X.~Lou, and C.~Choi, ``Interactive robotic grasping with
  attribute-guided disambiguation,'' \emph{arXiv preprint arXiv:2203.08037},
  2022.

\bibitem{zhang2021invigorate}
H.~Zhang, Y.~Lu, C.~Yu, D.~Hsu, X.~La, and N.~Zheng, ``Invigorate: Interactive
  visual grounding and grasping in clutter,'' \emph{arXiv preprint
  arXiv:2108.11092}, 2021.

\bibitem{goodwin2022semantically}
W.~Goodwin, S.~Vaze, I.~Havoutis, and I.~Posner, ``Semantically grounded object
  matching for robust robotic scene rearrangement,'' in \emph{2022
  International Conference on Robotics and Automation (ICRA)}.\hskip 1em plus
  0.5em minus 0.4em\relax IEEE, 2022, pp. 11\,138--11\,144.

\bibitem{fang2020graspnet}
H.-S. Fang, C.~Wang, M.~Gou, and C.~Lu, ``Graspnet-1billion: A large-scale
  benchmark for general object grasping,'' in \emph{Proceedings of the IEEE/CVF
  conference on computer vision and pattern recognition}, 2020, pp.
  11\,444--11\,453.

\bibitem{pinto2016supersizing}
L.~Pinto and A.~Gupta, ``Supersizing self-supervision: Learning to grasp from
  50k tries and 700 robot hours,'' in \emph{2016 IEEE international conference
  on robotics and automation (ICRA)}.\hskip 1em plus 0.5em minus 0.4em\relax
  IEEE, 2016, pp. 3406--3413.

\bibitem{mahler2017learning}
J.~Mahler and K.~Goldberg, ``Learning deep policies for robot bin picking by
  simulating robust grasping sequences,'' in \emph{Conference on robot
  learning}.\hskip 1em plus 0.5em minus 0.4em\relax PMLR, 2017, pp. 515--524.

\bibitem{kalashnikov2018scalable}
D.~Kalashnikov, A.~Irpan, P.~Pastor, J.~Ibarz, A.~Herzog, E.~Jang, D.~Quillen,
  E.~Holly, M.~Kalakrishnan, V.~Vanhoucke \emph{et~al.}, ``Scalable deep
  reinforcement learning for vision-based robotic manipulation,'' in
  \emph{Conference on Robot Learning}.\hskip 1em plus 0.5em minus 0.4em\relax
  PMLR, 2018, pp. 651--673.

\bibitem{ten2018using}
A.~Ten~Pas and R.~Platt, ``Using geometry to detect grasp poses in 3d point
  clouds,'' in \emph{Robotics research}.\hskip 1em plus 0.5em minus 0.4em\relax
  Springer, 2018, pp. 307--324.

\bibitem{mahler2017dex}
J.~Mahler, J.~Liang, S.~Niyaz, M.~Laskey, R.~Doan, X.~Liu, J.~A. Ojea, and
  K.~Goldberg, ``Dex-net 2.0: Deep learning to plan robust grasps with
  synthetic point clouds and analytic grasp metrics,'' \emph{arXiv preprint
  arXiv:1703.09312}, 2017.

\bibitem{sundermeyer2021contact}
M.~Sundermeyer, A.~Mousavian, R.~Triebel, and D.~Fox, ``Contact-graspnet:
  Efficient 6-dof grasp generation in cluttered scenes,'' in \emph{2021 IEEE
  International Conference on Robotics and Automation (ICRA)}.\hskip 1em plus
  0.5em minus 0.4em\relax IEEE, 2021, pp. 13\,438--13\,444.

\bibitem{wang2021graspness}
C.~Wang, H.-S. Fang, M.~Gou, H.~Fang, J.~Gao, and C.~Lu, ``Graspness discovery
  in clutters for fast and accurate grasp detection,'' in \emph{Proceedings of
  the IEEE/CVF International Conference on Computer Vision}, 2021, pp.
  15\,964--15\,973.

\bibitem{son2022grasping}
D.~Son, ``Grasping as inference: Reactive grasping in heavily cluttered
  environment,'' \emph{arXiv preprint arXiv:2205.13146}, 2022.

\bibitem{kiatos2019robust}
M.~Kiatos and S.~Malassiotis, ``Robust object grasping in clutter via
  singulation,'' in \emph{2019 International Conference on Robotics and
  Automation (ICRA)}.\hskip 1em plus 0.5em minus 0.4em\relax IEEE, 2019, pp.
  1596--1600.

\bibitem{kurenkov2020visuomotor}
A.~Kurenkov, J.~Taglic, R.~Kulkarni, M.~Dominguez-Kuhne, A.~Garg,
  R.~Mart{\'\i}n-Mart{\'\i}n, and S.~Savarese, ``Visuomotor mechanical search:
  Learning to retrieve target objects in clutter,'' in \emph{2020 IEEE/RSJ
  International Conference on Intelligent Robots and Systems (IROS)}.\hskip 1em
  plus 0.5em minus 0.4em\relax IEEE, 2020, pp. 8408--8414.

\bibitem{yang2020deep}
Y.~Yang, H.~Liang, and C.~Choi, ``A deep learning approach to grasping the
  invisible,'' \emph{IEEE Robotics and Automation Letters}, vol.~5, no.~2, pp.
  2232--2239, 2020.

\bibitem{xu2021efficient}
K.~Xu, H.~Yu, Q.~Lai, Y.~Wang, and R.~Xiong, ``Efficient learning of
  goal-oriented push-grasping synergy in clutter,'' \emph{IEEE Robotics and
  Automation Letters}, vol.~6, no.~4, pp. 6337--6344, 2021.

\bibitem{huang2021visual}
B.~Huang, S.~D. Han, J.~Yu, and A.~Boularias, ``Visual foresight trees for
  object retrieval from clutter with nonprehensile rearrangement,'' \emph{IEEE
  Robotics and Automation Letters}, vol.~7, no.~1, pp. 231--238, 2021.

\bibitem{stepputtis2020language}
S.~Stepputtis, J.~Campbell, M.~Phielipp, S.~Lee, C.~Baral, and H.~Ben~Amor,
  ``Language-conditioned imitation learning for robot manipulation tasks,''
  \emph{Advances in Neural Information Processing Systems}, vol.~33, pp.
  13\,139--13\,150, 2020.

\bibitem{rao2018learning}
A.~B. Rao, K.~Krishnan, and H.~He, ``Learning robotic grasping strategy based
  on natural-language object descriptions,'' in \emph{2018 IEEE/RSJ
  International Conference on Intelligent Robots and Systems (IROS)}.\hskip 1em
  plus 0.5em minus 0.4em\relax IEEE, 2018, pp. 882--887.

\bibitem{ito2022integrated}
H.~Ito, H.~Ichiwara, K.~Yamamoto, H.~Mori, and T.~Ogata, ``Integrated learning
  of robot motion and sentences: Real-time prediction of grasping motion and
  attention based on language instructions,'' in \emph{2022 International
  Conference on Robotics and Automation (ICRA)}.\hskip 1em plus 0.5em minus
  0.4em\relax IEEE, 2022, pp. 5404--5410.

\bibitem{chen2021joint}
Y.~Chen, R.~Xu, Y.~Lin, and P.~A. Vela, ``A joint network for grasp detection
  conditioned on natural language commands,'' in \emph{2021 IEEE International
  Conference on Robotics and Automation (ICRA)}.\hskip 1em plus 0.5em minus
  0.4em\relax IEEE, 2021, pp. 4576--4582.

\bibitem{misra2016tell}
D.~K. Misra, J.~Sung, K.~Lee, and A.~Saxena, ``Tell me dave: Context-sensitive
  grounding of natural language to manipulation instructions,'' \emph{The
  International Journal of Robotics Research}, vol.~35, no. 1-3, pp. 281--300,
  2016.

\bibitem{stengel2022guiding}
E.~Stengel-Eskin, A.~Hundt, Z.~He, A.~Murali, N.~Gopalan, M.~Gombolay, and
  G.~Hager, ``Guiding multi-step rearrangement tasks with natural language
  instructions,'' in \emph{Conference on Robot Learning}.\hskip 1em plus 0.5em
  minus 0.4em\relax PMLR, 2022, pp. 1486--1501.

\bibitem{ahn2018interactive}
H.~Ahn, S.~Choi, N.~Kim, G.~Cha, and S.~Oh, ``Interactive text2pickup networks
  for natural language-based human--robot collaboration,'' \emph{IEEE Robotics
  and Automation Letters}, vol.~3, no.~4, pp. 3308--3315, 2018.

\bibitem{yu2016modeling}
L.~Yu, P.~Poirson, S.~Yang, A.~C. Berg, and T.~L. Berg, ``Modeling context in
  referring expressions,'' in \emph{European Conference on Computer
  Vision}.\hskip 1em plus 0.5em minus 0.4em\relax Springer, 2016, pp. 69--85.

\bibitem{vaswani2017attention}
A.~Vaswani, N.~Shazeer, N.~Parmar, J.~Uszkoreit, L.~Jones, A.~N. Gomez,
  {\L}.~Kaiser, and I.~Polosukhin, ``Attention is all you need,''
  \emph{Advances in neural information processing systems}, vol.~30, 2017.

\bibitem{mildenhall2020nerf}
B.~Mildenhall, P.~P. Srinivasan, M.~Tancik, J.~T. Barron, R.~Ramamoorthi, and
  R.~Ng, ``Nerf: Representing scenes as neural radiance fields for view
  synthesis,'' in \emph{Computer Vision--ECCV 2020: 16th European Conference,
  Glasgow, UK, August 23--28, 2020, Proceedings, Part I}, 2020, pp. 405--421.

\bibitem{haarnoja2018soft}
T.~Haarnoja, A.~Zhou, K.~Hartikainen, G.~Tucker, S.~Ha, J.~Tan, V.~Kumar,
  H.~Zhu, A.~Gupta, P.~Abbeel \emph{et~al.}, ``Soft actor-critic algorithms and
  applications,'' \emph{arXiv preprint arXiv:1812.05905}, 2018.

\bibitem{christodoulou2019soft}
P.~Christodoulou, ``Soft actor-critic for discrete action settings,''
  \emph{arXiv preprint arXiv:1910.07207}, 2019.

\bibitem{appendix_arxiv}
K.~Xu, ``Paper with appendix,'' \url{https://arxiv.org/abs/2302.12610}, 2021.

\bibitem{coumans2021}
E.~Coumans and Y.~Bai, ``Pybullet, a python module for physics simulation for
  games, robotics and machine learning,'' \url{http://pybullet.org},
  2016--2021.

\bibitem{he2016deep}
K.~He, X.~Zhang, S.~Ren, and J.~Sun, ``Deep residual learning for image
  recognition,'' in \emph{Proceedings of the IEEE conference on computer vision
  and pattern recognition}, 2016, pp. 770--778.

\bibitem{liu2021ocrtoc}
Z.~Liu, W.~Liu, Y.~Qin, F.~Xiang, M.~Gou, S.~Xin, M.~A. Roa, B.~Calli, H.~Su,
  Y.~Sun \emph{et~al.}, ``Ocrtoc: A cloud-based competition and benchmark for
  robotic grasping and manipulation,'' \emph{IEEE Robotics and Automation
  Letters}, vol.~7, no.~1, pp. 486--493, 2021.

\bibitem{perez2018film}
E.~Perez, F.~Strub, H.~De~Vries, V.~Dumoulin, and A.~Courville, ``Film: Visual
  reasoning with a general conditioning layer,'' in \emph{Proceedings of the
  AAAI Conference on Artificial Intelligence}, vol.~32, no.~1, 2018.

\end{thebibliography}


\cleardoublepage
\appendix
\subsection{Details about CLIP Grounding Prior}
\label{clip-grounding}
By calculating the cosine similarity between $N$ box visual feature and the text feature, we can ground $N$ visual-language pairs into $N$ probabilities, each of which represents a matching score of the corresponding object bounding box against the language instruction. We obtain 3D positions $(x, y, z)$ of bounding box centers by transforming pixel coordinates to world coordinates with camera extrinsics for $(x, y)$ and getting the depth of the center pixel of depth image for $z$. Leveraging the 3D positions of the bounding box centers, 3D distance between each grasp and each bounding box can be calculated. If the distance of a grasp and a bounding box is smaller than a distance threshold $d=0.05$ (a hyper parameter referring to \cite{liu2021ocrtoc}, which uses graspnet \cite{fang2020graspnet} to generate grasps and assigns grasps to objects), then the grasp is assigned to that object, which results in a $N\times K$ box-grasp mapping matrix. By multiplying CLIP probabilities and the box-grasping mapping matrix, we can get $K$ mapped probabilities, which can be regarded as a grounding prior for the policy. Then a guided loss ({\it i.e.} Kullback-Leibler divergence) can be added for the policy training.

\subsection{More Training Details}
Our training object models are from GraspNet-1Billion \cite{fang2020graspnet}. For simplicity, we generate object bounding boxes from the mask image in Pybullet \cite{coumans2021}, of which each pixel donates the index of the object visualized in the camera. Specifically, the bounding box of pixels with the same index corresponds to an object-centric representation. And the bounding box whose size is smaller than $15 \times 15$ will be abandoned. Note that this representation is not a ground-truth bounding box, but considering object occlusion. During the training stage, we randomly sample a language template and a keyword to form a complete language instruction. And the probability of the keyword sampling is 0.4, 0.2, 0.2, 0.2 for the above four types. For the keyword type of general label and function, which is regarded to form the difficult scenarios, we will drop two target objects in the scene, and successfully catching one is regarded as finishing the task. For the rest type of keywords, one target object is dropped into the workspace, and grasping the target object means task completion. 

3D positions of box centers are obtained as described in Appendix. \ref{clip-grounding}. And before feeding into the network, the positions are projected into a nonlinear space by positional encoding as in \cite{mildenhall2020nerf}. We adopt the transformer architecture of the text encoder in \cite{radford2021learning}, and conduct multi-head attention with different query, key and value. Parameters of the cross transformer includes width of 512, head of 8 and layer of 1.



We train our method with discrete SAC \cite{christodoulou2019soft}. The temperature parameter $\alpha$ is initialized as 0.2 with automatic entropy tuning. We define the policy as $\pi(a_t|s_t)$, where $s_t$ donates the cross-attention state at time $t$, and $a_t$ represents the grasp action at time $t$, The critic outputs Q function $Q(s_t, a_t)$, then the loss function of actor, critic and $\alpha$ can be formulated as Eqn. \ref{eq-1}-\ref{eq-3}. And the loss of the cross-attention module is the sum of policy loss and critic loss. All the networks are trained with Adam optimizer using fixed learning rates $3\times 10^{-4}$. Our future discount $\gamma$ is set as a constant at 0.99.
\begin{equation}
\label{eq-1}
\mathcal{L}_\pi(\theta)=\pi_\theta(a_t|s_t)^T\left(-Q_\theta(s_t, a_t)+\alpha \log \pi_\theta(a_t|s_t)\right)
\end{equation}
\begin{equation}
\label{eq-2}
\begin{gathered}
\mathcal{L}_Q(\theta)=\left(Q_\theta(s_t, a_t)-y\left(r_t, s_{t\sim t+1}, a_{t\sim t+1}\right)\right)^2 \\
y\left(r_t, s_{t\sim t+1}, a_{t\sim t+1}\right)=r_t+\gamma\left(\pi_\theta(a_t|s_t)^T V_\theta(s_{t+1}, a_{t+1})\right) \\
V_\theta(s_{t+1}, a_{t+1})=Q_\theta\left(s_{t+1}, a_{t+1}\right)-\alpha \log \pi_\theta\left(a_{t+1}| s_{t+1}\right)
\end{gathered}
\end{equation}
\begin{equation}
\label{eq-3}
\mathcal{L}_\alpha(\alpha)=\pi_\theta(a_t|s_t)^T\left(-\alpha \log \pi_\theta(a_t|s_t)+\bar{H}\right)
\end{equation}
where $r_t$ is the reward at step $t$, $\theta$ donates the network parameters, and $\bar{H}$ is the target entropy.

\subsection{Details about Object Detection Model}
We deploy a pre-trained open-set detection model in our real-world experiments. The detection model is trained with data from GraspNet-1Billion \cite{fang2020graspnet} with $mAP=70.70$ for known objects and $mAP=34.53$ for unknown objects.

\subsection{Ablation Studies of Architecture Design}
In this section, we compare our methods with two variants to verify the design of our joint modeling. Compared results are shown in Table \ref{table:5}. {\bf Position as Key} is a variant which directly uses the 3D positions of bounding boxes as the keys of cross attention, while {\bf FiLM Fusion} is a variant to fuse the cross attention feature of visual and action with the language feature by FiLM \cite{perez2018film}, which is a fusion technique for visual-language reasoning. It is obvious that the performance of both variants are unsatisfactory. {\bf Position as Key} tries to simply combine geometry information ({\it i.e.} grasp pose) and visual-language information by the 3D positions of bounding boxes, which might overfit to seen scenarios. {\bf FiLM Fusion} fails possibly due to the fact that the visual feature and language feature has been aligned into one space by CLIP, but FiLM additionally introduces a inductive bias, which makes the fusion feature out of the distribution of CLIP domain.

\begin{table}[t]
\scriptsize
\caption{{ABLATION RESULTS FOR ARCHITECTURE DESIGN}}
\label{table:5}
\vspace{-0.4cm}
\begin{center}
\begin{tabular}{p{2.4cm}|p{0.8cm}|p{0.8cm}|p{0.8cm}|p{0.8cm}}
\hline
\makecell[c]{Method} & \multicolumn{2}{c|}{Task Success} & \multicolumn{2}{c}{Motion Number} \\
\hline
\makecell[c]{Arrangement} & \makecell[c]{seen} & \makecell[c]{unseen} & \makecell[c]{seen}  & \makecell[c]{unseen} \\
\hline
\makecell[c]{Position as Key} & \makecell[c]{57.3} & \makecell[c]{48.0} & \makecell[c]{4.65} & \makecell[c]{4.23}\\
\makecell[c]{FiLM Fusion} & \makecell[c]{58.7} & \makecell[c]{38.7} & \makecell[c]{4.28} & \makecell[c]{4.67}\\
\makecell[c]{Ours} & \makecell[c]{{\color{red} \bf74.3}} & \makecell[c]{{\color{red} \bf78.7}} & \makecell[c]{{\color{red} \bf4.11}} & \makecell[c]{{\color{red} \bf3.98}}\\
\hline
\end{tabular}
\end{center}
\vspace{-0.2cm}
\end{table}

\subsection{Example Action Sequences in Real-world}
\begin{figure}[t]
  \centering
  \includegraphics[width=\linewidth]{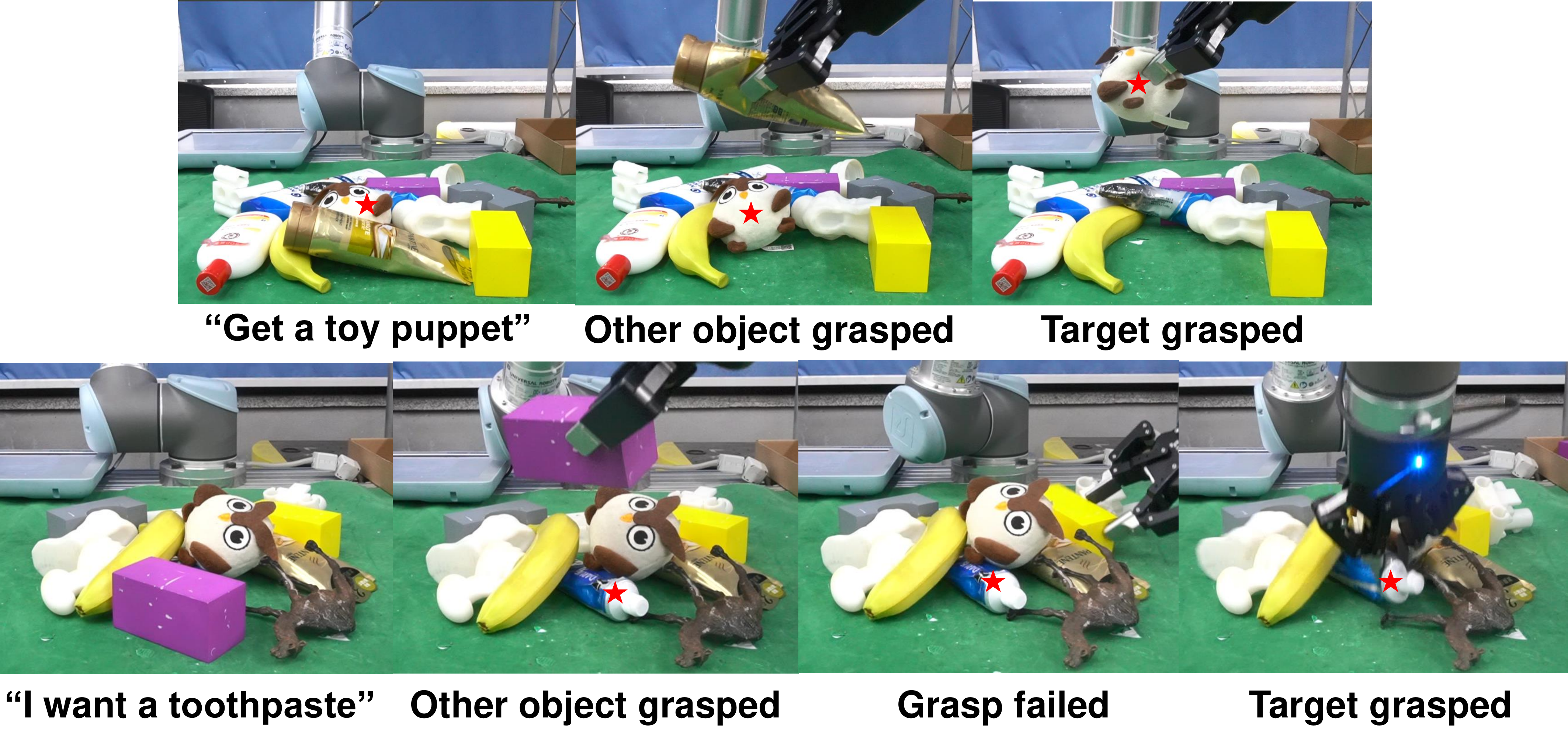}
  \vspace{-0.6cm}
  \caption{Example testing sequences in real world. Target objects are labeled with stars.}
  \label{fig:real_sequence}
  \vspace{-0.5cm}
\end{figure}

Fig. \ref{fig:real_sequence} shows two example testing sequences in real world. In these two cases, the target objects ({\it i.e.} toy puppet and toothpaste) are unseen during training in simulation. For the first case, the robot is able to grasp away the surrounding object to make space for the target object grasping. For the second case, the target object is occluded by other objects, and fails to extracted by the detection module at the beginning. Instead, the target object is partially included in other objects' bounding box. We observe that in this situation, our policy can also grasp away the near objects and finally successfully grasp the target, which demonstrate that our policy might capture the target information in the surrounding bounding boxes, which guides the exploration of the policy. 

In real-world experiments, failure happens mainly when the detection model detects a wrong target object, or the graspnet fails to detect a successful grasp of the target object.

\subsection{Limitations and Future Works}

In this work, we focus on the grasping task in language-conditioned manipulation. Our method is trained and tested with language instructions containing only one concept. However, we believe compound concepts such as "a red and round object" can be integrated into this work by additional finetuning. Additionally whether grasping the target object is decided by human. In future work, we will consider distinguishing the final state by the model itself. Also, more complex tasks, such as pick-and-place, or tool use, can be extended in the future.

\end{document}